# Compliance error compensation technique for parallel robots composed of non-perfect serial chains

Alexandr Klimchik [a,b],[1] Anatol Pashkevich [a,b], Damien Chablat [b], Geir Hovland [c]

[a] *Ecole des Mines de Nantes, 4 rue Alfred-Kastler, Nantes 44307, France*
[b] *Institut de Recherches en Communications et en Cybernétique de Nantes, UMR CNRS 6597, 1 rue de la Noe, 44321 Nantes, France*
[c] *University of Agder, Gimlemoen 25A, Kristiansand, Norway,*

*Abstract*

*The paper presents the compliance errors compensation technique for over-constrained parallel manipulators under external and internal loadings. This technique is based on the non-linear stiffness modeling which is able to take into account the influence of non-perfect geometry of serial chains caused by manufacturing errors. Within the developed technique, the deviation compensation reduces to an adjustment of a target trajectory that is modified in the off-line mode. The advantages and practical significance of the proposed technique are illustrated by an example that deals with groove milling by the Orthoglide manipulator that considers different locations of the workpiece. It is also demonstrated that the impact of the compliance errors and the errors caused by inaccuracy in serial chains cannot be taken into account using the superposition principle.*

## 1 Introduction

In machining applications, robot accuracy depends on a numbers of factors [1]. The most essential of them are related to manufacturing tolerances leading to geometrical parameters deviation with respect to their nominal values (geometrical errors) as well as to the end-effector deflections caused by the cutting forces and torques (compliance errors). Usually, in applications where the external forces/torques applied to the end-effector are relatively small, the prime source of manipulator inaccuracy are the *geometrical errors* [2, 3]. These errors are associated with differences between nominal and actual values of the link/joint parameters. Typical examples of them are the differences between the nominal and the actual lengths of links, differences between zero values of actuator coordinates in real robot and mathematical model embedded in controller (joint offsets); they can also be induced by non-perfect assembling of different elements and arise in shifting and/or rotation of the frames associated with different components, which normally are assumed to be matched and aligned. It is clear that the geometrical errors do not depend on the manipulator configuration, while their effect on the position accuracy is dependent on it. At present, there exists a number of sophisticated calibration techniques that are able to identify differences between actual and nominal geometrical parameters [4-9]. Consequently, these types of errors can be efficiently compensated by either by adjusting the controller input (i.e. the target point coordinates) or by straightforward modification of geometrical model parameters used in the robot controller.

In some other cases, the geometrical errors may be dominated by *non-geometrical* ones that are caused by influence of a number of factors [10-12]. For instance, the forces/torques associated with the tool-workpiece interaction in the technological process may cause essential deformations of the manipulator components (compliance errors) [13]. Also, the environmental conditions (temperature, atmospheric pressure and others) may affect physical properties of manipulator components and lead to undesirable changes in link dimensions. It is worth mentioning that, in the regular service conditions, the *compliance errors* are the most significant ones. Generally, they depend on two main factors: (i) the stiffness of the robotic manipulator and (ii) loading applied to it. Similar to the geometrical ones, the compliance errors highly depend on the manipulator configuration and essentially differ throughout the workspace. Their influence is particularly important for heavy robots and for manipulators with low stiffness. For example, the cutting forces/torques from the technological process may induce significant deformations, which are not negligible in the precise machining. In this case, the influence of the compliance errors on the robot positional accuracy can be even higher than the geometrical ones. This issue is very important for the designers of parallel manipulators, who often are looking for a compromise between the manipulator stiffness and its dynamic capabilities [14].

Thus, this paper focuses on the compliance error compensation that is able to take into account both conventional error sources (compliance errors caused by external and internal forces/torques) and errors caused by assembling of non-perfect over-constrained parallel mechanisms. To address these problems, the remainder of the paper is organized as follows: Section 2 presents review on the compliance error compensation methods, Section 3 provides required background for the stiffness modeling, Section 4 proposes the non-linear compliance errors compensation technique, in Section 5 the efficiency of the developed technique is illustrated by a simulation example that deals with the groove milling using Orthoglide manipulator, and, finally, Section 6 summarizes the main results of the paper.

---

[1] Corresponding author. Tel. +33 251 85 83 17; fax. +33 251 85 83 49; E-mail address: alexandr.klimchik@mines-nantes.fr (A. Klimchik).

## 2 Problem of compliance error compensation

In many robotic applications such as machining, grinding, trimming etc., the interaction between the workpiece and the end-effector causes essential deflections that significantly decrease the processing accuracy and quality of the final product. To overcome this difficulty, it is possible to modify either control algorithm or the prescribed trajectory, which is used as the reference input for a control system [15]. This paper focuses on the second approach that is considered to be more realistic in the practice. In contrast to the previous works, the proposed compliance error compensation technique is based on the non-linear stiffness model of the manipulator that is able to take into account significant external loading [16].

Usually, the problem of the robot error compensation can be solved in two ways that differ in degree of modification of the robot control software:

(a) by *modification of the manipulator model* (Figure 1a) which better suits to the real manipulator and is used by the robot controller (in simple case, it can be limited by tuning of the nominal manipulator model, but may also involve essential model enhancement by introducing additional parameters, if it is allowed by a robot manufacturer);

(b) by *modification of the robot control program* (Figure 1b) that defines the prescribed trajectory in Cartesian space (here, using relevant error model, the input trajectory is generated in such way that under the loading the output trajectory coincides with the desired one).

It is clear that the first approach can be implemented in on-line mode, while the second one requires preliminary off-line computations. It is worth mentioning that the stiffness models being used in this work are suitable for both of these approaches. But in practice it is rather unrealistic to include the stiffness model in a commercial industrial controller where all transformations between the joint and Cartesian coordinates are based on the manipulator geometrical model. In contrast, the off-line error compensation, based on the second approach, is attractive for industrial applications.

For the *geometrical errors*, relevant compensation techniques are already well developed. A comprehensive review of related works is given in [2, 9, 17]. Here, if the main error sources are concentrated in the link length or in the joint offsets, the compensation is reduced to straightforward modification of the manipulator parameters in the robot controller. Otherwise, if there are any geometrical error sources that are not presented in the nominal inverse/direct kinematics, relevant modification of the controller input is required. In this case, it is possible to use (in off-line mode) either extended geometrical model with additional parameters or simply a non-linear function that describes the error distribution throughout the workspace. Examples of such a function are given in [18, 19] where the neural network technique is employed. In this framework, it is assumed that the geometrical errors are less essential compared to the non-geometrical ones caused by the interaction between the machining tool and workpiece. So, the main attention will be paid to the *compliance errors* and their compensation techniques.

For the *compliance errors*, the compensation technique must rely on two components. The first of them describes distribution of the stiffness properties throughout the workspace and is defined by the stiffness matrix as a function of the joint coordinates or the Cartesian location [16]. The second component describes the forces/torques acting on the end-effector while the manipulator is performing its manufacturing task (manipulator *loading*). In this work, it is assumed that the second component is given and can be obtained either from the dedicated technological process model or by direct measurements using the force/torque sensor integrated into the end-effector. However it is necessary to take into account that the force sensors introduce additional undesirable compliance which has a direct affect on the position accuracy [20, 21].

The *stiffness matrix* required for the compliance errors compensation highly depends on the robot configuration and essentially varies throughout the workspace. Hypothetically, it can be also approximated by a neural network, similar to the geometrical error compensation mentioned above. However, this approach is not attractive practically, so it is more convenient to compute the stiffness matrix using specially developed expressions and algorithms.

From general point of view, full-scale compensation of the compliance errors requires essential revision of the manipulator model embedded in the robot controller. In fact, instead of conventional geometrical model that provides inverse/direct coordinate transformations from the joint to Cartesian spaces and vice versa, here it is necessary to employ the so-called *kinetostatic model* [22]. The later defines the mapping between the joint and Cartesian spaces taking into account deflections caused by external forces/torques applied to the manipulator end-effector. It is essentially more complicated than the geometrical model and requires rather intensive computations that are presented in Section 3.

*(a) Modification of the manipulator model*

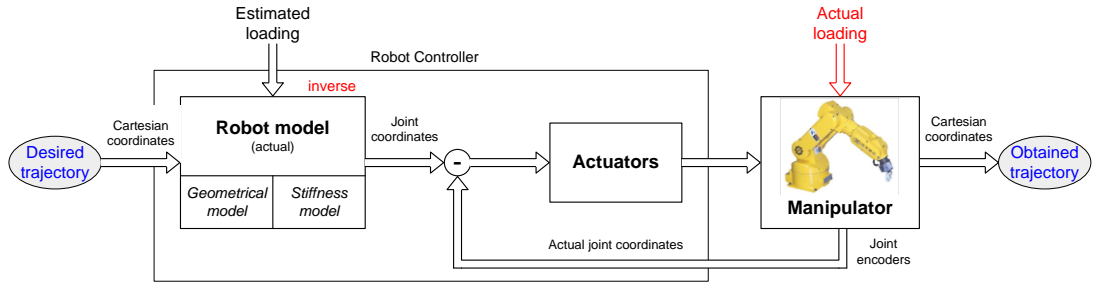

*(b) Modification of the target trajectory*

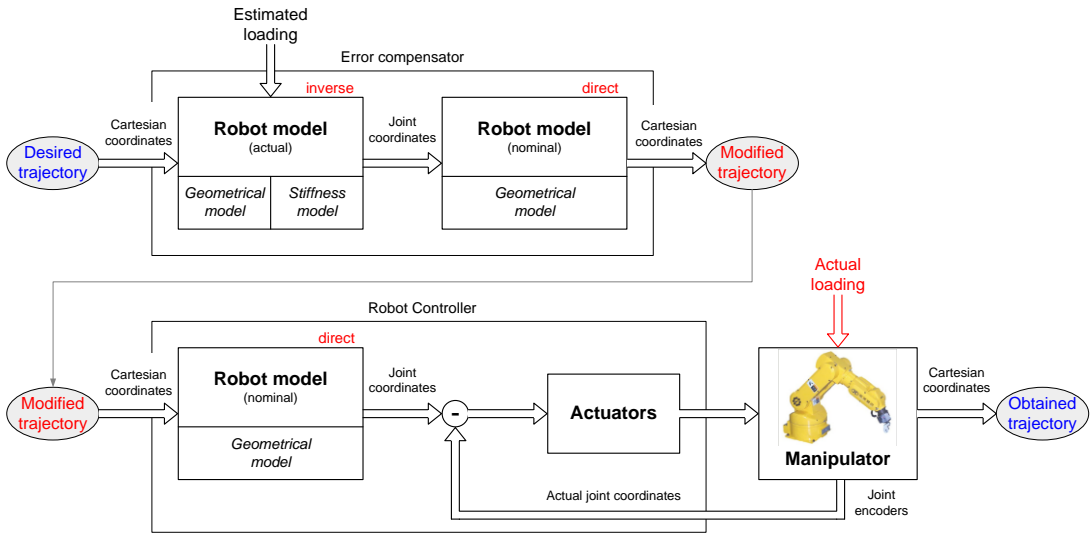

**Figure 1**    Robot error compensation methods

If the compliance errors are relatively small, composition of conventional geometrical model and the stiffness matrix give rather accurate approximation of the modified mapping from the joint to Cartesian space. In this case, for the first compensation scheme (see Figure 1a), the kinetostatic model can be easily implemented online if there is access to the control software modification. Otherwise, the second scheme (see Figure 1b) can be easily applied. Moreover, with regard to the robot-based machining, there is a solution that does not require force/torque measurements or computations [15, 23, 24]. Its basic idea is presented in Figure 2, where at the first stage it is performed, the machining experiment gives a trajectory corrupted by the compliance errors. Then, the difference between the desired and the obtained trajectories is evaluated via appropriate measurements, which give the compliance errors along the path. Using this data and assuming that the stiffness model is linear, the target trajectory for the robot controller is modified by applying the "mirror" technique (where corresponding points of the corrupted and target trajectories are symmetrical with respect to the relevant points of the desired trajectory). In order to improve accuracy in [25] it was proposed to perform the machining experiments several times. Starting from the second experiment the target trajectory for the robot controller has been modified by applying the "mirror" technique for the measurements obtained during previous experiments. An evident advantage of this technique is its applicability to the compensation of all types of the robot errors, including geometrical and compliance ones. However, this approach is only suitable for the large-scale production where the manufacturing task and the workpiece location remains the same. Eastwood and Webb [26] proposed polar compensation methodology for gravitational deflection compensation for hybrid parallel kinematic machines. In some other works [26, 27] the problems of geometrical and compliance errors compensation have been considered simultaneously but these techniques cannot be applied to robot-based machining directly.

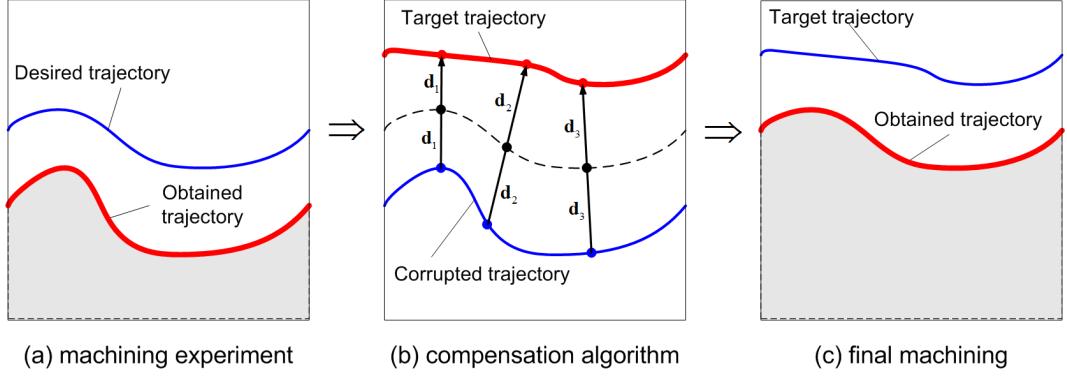

**Figure 2**   Method of symmetrical trajectory for compensation of the compliance errors

Hence, to be applied to the robotic-based machining, the existing compliance errors compensation techniques should be essentially revised to take into account essential forces and torques as well as some other important error sources (inaccuracy in serial chains, for instance). This problem is in the scope of this work.

## 3   Stiffness modeling background

The compliance error compensation technique developed in this work is based on our previous results [16], which provide a nonlinear stiffness modeling technique for the manipulators with passive joints. This approach is based on the Virtual Joint Modeling (VJM) method, which extended the conventional rigid-body model of the robotic manipulator by adding virtual springs that take into account elastostatic properties of links and joints. The method proposed by Salisbury in 1980 [29] has been modified several times [29-33] and at present it is the most popular stiffness analysis method in robotics. Let us summarise here the main expressions that will be used in the nonlinear compliance error compensation technique.

It is assumed that the end-points of all kinematic chains are aligned and matched in the same target point $t_0$, which corresponds to the desired end-platform location. This point is assumed to be known and allows us to compute, from the inverse kinematic model, the actuator and passive joint coordinates defining nominal configurations of the chains $(q_{0i}, \theta_{0i})$. Static equilibrium for serial chain that corresponds to applied loading can be obtained using the principle of virtual displacements and computed using the following iterative procedure

$$\begin{bmatrix} \mathbf{F}'_i \\ \mathbf{q}'_i \\ \mathbf{\theta}'_i \end{bmatrix} = \begin{bmatrix} \mathbf{0} & \mathbf{J}_{qi}(\mathbf{q}_i,\mathbf{\theta}_i) & \mathbf{J}_{\theta i}(\mathbf{q}_i,\mathbf{\theta}_i) \\ \mathbf{J}_{qi}^T(\mathbf{q}_i,\mathbf{\theta}_i) & \mathbf{0} & \mathbf{0} \\ \mathbf{J}_{\theta i}^T(\mathbf{q}_i,\mathbf{\theta}_i) & \mathbf{0} & -\mathbf{K}_{\theta i} \end{bmatrix}^{-1} \cdot \begin{bmatrix} \mathbf{t} - \mathbf{g}_i(\mathbf{q}_i,\mathbf{\theta}_i) + \mathbf{J}_{qi}(\mathbf{q}_i,\mathbf{\theta}_i)\cdot\mathbf{q}_i + \mathbf{J}_{\theta i}(\mathbf{q}_i,\mathbf{\theta}_i)\cdot\mathbf{\theta}_i \\ \mathbf{0} \\ -\mathbf{K}_{\theta i}\cdot\mathbf{\theta}_{0i} \end{bmatrix} \qquad (1)$$

where the subscript "(i)" denotes the serial chain number, the prime define new static equilibrium configuration that should be used in the next iteration as a current one, $\mathbf{F}$ is external loading applied to the end-point of kinematic chain, joint coordinates $(\mathbf{q},\mathbf{\theta})$ define serial chain configuration that corresponds the applied loading $\mathbf{F}$, $\mathbf{J}_q(\mathbf{q},\mathbf{\theta})$ and $\mathbf{J}_\theta(\mathbf{q},\mathbf{\theta})$ are Jacobian matrices with respect to passive $\mathbf{q}$ and virtual $\mathbf{\theta}$ joint coordinates respectively computed for current serial chain configuration, matrix $\mathbf{K}_\theta$ is stiffness matrix of serial chain in the joint coordinates, vector $\mathbf{t}$ defines end-effector location under the loading. Function $\mathbf{g}(\mathbf{q},\mathbf{\theta})$ defines geometry of serial chain, $\mathbf{\theta}_0$ is preloading in virtual joints. The iterative scheme (1) allows us to obtain the relation between the elastic deformations in joints and corresponding external force/torque applied to the end-effector. It is based on sequential computations of loaded equilibriums (and relevant force/torque) for the displacement of the manipulator end-point with respect to its unloaded location. So, partial non-linear force-deflection relations corresponding to the target point $t_0$ can be presented in the form

$$\mathbf{F}_i = f_i(\mathbf{t} \mid \mathbf{t}_0) \qquad (2)$$

Then, in accordance with the superposition principle, the non-linear force-deflection relation for the whole parallel manipulator can be found by straightforward summation of all partial forces $\mathbf{F}_i$, i.e.

$$\mathbf{F} = \sum_{i=1}^{m} f_i(\mathbf{t} \mid \mathbf{t}_0) \qquad (3)$$

where $\mathbf{F}$ denotes the total external loading applied to the end-platform. As a result, corresponding curves can be obtained by multiple repetition of the above described procedures for different values of the end-platform location $\mathbf{t}$.

Furthermore, for each given $\mathbf{t}$, the Cartesian stiffness matrices $\mathbf{K}_C^{(i)}$ of all kinematic chains, that correspond to the loading configuration can be computed using the following expression

$$\mathbf{K}_C^{(i)} = \mathbf{K}_C^{0(F)} - \mathbf{K}_C^{q(F)} \tag{4}$$

where the first term $\mathbf{K}_C^{0(F)} = (\mathbf{J}_\theta \cdot \mathbf{k}_\theta^F \cdot \mathbf{J}_\theta^T)^{-1}$ corresponds exactly to the classical formula defining stiffness of the kinematic chain without passive joints in the loaded mode [34, 35] and the second term $\mathbf{K}_C^{q(F)}$ takes into account influence of passive joints and can be computed as

$$\mathbf{K}_C^{q(F)} = -\mathbf{K}_C^{0(F)} \cdot \mathbf{J}_q^{(F)} \cdot \left( \mathbf{H}_{qq}^F + \mathbf{H}_{q\theta}^F \cdot \mathbf{k}_\theta^F \cdot \mathbf{H}_{\theta q}^F - \mathbf{J}_q^{(F)T} \cdot \mathbf{K}_C^{0(F)} \cdot \mathbf{J}_q^{(F)} \right)^{-1} \cdot \mathbf{J}_q^{(F)T} \cdot \mathbf{K}_C^{0(F)} \tag{5}$$

where $\mathbf{J}_q^{(F)} = \mathbf{J}_q + \mathbf{J}_\theta \cdot \mathbf{k}_\theta^F \cdot \mathbf{H}_{\theta q}^F$.

Finally, this allows us to compute the Cartesian stiffness matrix $\mathbf{K}_C$ of the whole parallel robot as a sum

$$\mathbf{K}_C = \sum_{i=1}^{m} \mathbf{K}_C^{(i)} \tag{6}$$

This approach allows us to obtain the non-linear force-deflection relation for a parallel manipulator in the loaded mode as well as to compute Cartesian stiffness matrices for any given target point $\mathbf{t}_0$ and given the end-point location $\mathbf{t}$. However, it cannot be applied for the compliance errors compensation directly. Thus a dedicated technique, that is considered in this paper, is required.

## 4 Nonlinear technique for compliance error compensation

In industrial robotic controllers, the manipulator motions are usually generated using the inverse kinematic model that allows us to compute the input signals for actuators $\boldsymbol{\rho}_0$ corresponding to the desired end-effector location $\mathbf{t}_0$, which is assigned assuming that the compliance errors are negligible. However, if the external loading is essential, the kinematic control becomes non-applicable because of changes in the end-platform location. It can be computed from the non-linear compliance model as

$$\mathbf{t}_F = f^{-1}(\mathbf{F} \mid \mathbf{t}_0) \tag{7}$$

where the subscript 'F' refers to the loaded mode.

To compensate this undesirable end-platform displacement from $\mathbf{t}_0$ to $\mathbf{t}_F$, the target point should be modified in such a way that, under the loading $\mathbf{F}$, the end-platform is located in the desired point $\mathbf{t}_0$. This requirement can be expressed using the stiffness model in the following way

$$\mathbf{F} = f\left(\mathbf{t}_0 \mid \mathbf{t}_0^{(F)}\right) \tag{8}$$

where $\mathbf{t}_0^{(F)}$ denotes the modified target location. Hence, the problem is reduced to the solution of the nonlinear equation (8) for $\mathbf{t}_0^{(F)}$, while $\mathbf{F}$ and $\mathbf{t}_0$ are assumed to be given. It is worth mentioning that this equation completely differs from the equation $\mathbf{F} = f(\mathbf{t} \mid \mathbf{t}_0)$, where the unknown variable is $\mathbf{t}$. It means that here the compliance model does not allow us to compute the modified target point $\mathbf{t}_0^{(F)}$ directly, while the linear compensation technique directly operates with Cartesian compliance matrix [10, 23].

To solve equation (8) for $\mathbf{t}_0^{(F)}$, the Newton-Raphson technique can be applied. It yields the following iterative scheme

$$\mathbf{t}_0^{(F)\prime} = \mathbf{t}_0^{(F)} + \mathbf{K}_{t.p.}^{-1}(\mathbf{t}_0 \mid \mathbf{t}_0^{(F)}) \cdot \left(\mathbf{F} - f(\mathbf{t}_0 \mid \mathbf{t}_0^{(F)})\right) \tag{9}$$

where the prime corresponds to the next iteration and $\mathbf{K}_{t.p.}(\mathbf{t}_0 \mid \mathbf{t}_0^{(F)})$ is Cartesian stiffness matrix computed with respect to the second argument of the stiffness model $\mathbf{F} = f(\mathbf{t} \mid \mathbf{t}_0)$

$$\mathbf{K}_{t.p.}(\mathbf{t} \mid \mathbf{t}_0) = \frac{\partial f(\mathbf{t} \mid \mathbf{t}_0)}{\partial \mathbf{t}_0} \tag{10}$$

This argument $\mathbf{t}_0$ can be interpreted as the target point. Here, the location $\mathbf{t}_0$ can also be used as the initial value of $\mathbf{t}_0^{(F)}$. The stopping criterion can be expressed as

$$\left\| \mathbf{F} - f\left(\mathbf{t}_0 \mid \mathbf{t}_0^{(F)}\right) \right\| < \varepsilon_F \tag{11}$$

where $\varepsilon_F$ is the desired accuracy.

To overcome computational difficulties related to the evaluation of the matrix $\mathbf{K}_{t.p.}(\mathbf{t}_0 | \mathbf{t}_0^{(F)})$, it is possible to use its simple approximation that does not change from iteration to iteration. In particular, assuming that $\mathbf{t}$ and $\mathbf{t}_0$ are close enough and the stiffness properties do not vary substantially in their neighborhood, the stiffness model can be approximated by a linear expression $\mathbf{F} = \mathbf{K}_C(\mathbf{t} - \mathbf{t}_0)$, which gives $\mathbf{K}_{t.p.} = -\mathbf{K}_C$. Hence, the iterative scheme (9) can be modified as

$$\mathbf{t}_0^{(F)\prime} = \mathbf{t}_0^{(F)} - \alpha \cdot \mathbf{K}_C^{-1}(\mathbf{t}_0 | \mathbf{t}_0^{(F)}) \cdot \left( \mathbf{F} - f(\mathbf{t}_0 | \mathbf{t}_0^{(F)}) \right) \tag{12}$$

where $\alpha \in (0,1)$ is the scalar parameter ensuring the convergence. Using the non-linear compliance model (7), this idea can also be implemented in an iterative algorithm

$$\mathbf{t}_0^{(F)\prime} = \mathbf{t}_0^{(F)} + \alpha \cdot \left( \mathbf{t}_0 - f^{-1}(\mathbf{F} | \mathbf{t}_0^{(F)}) \right) \tag{13}$$

which does not include stiffness matrices $\mathbf{K}_C$ or $\mathbf{K}_{t.p.}$. Obviously, this is the most computationally convenient solution and it will be used in the next section.

It should be mentioned that the considered case deals with a perfect parallel manipulator where end-points of all kinematic chains are aligned and matched. However, in practice, kinematic chains may include some errors that do not allow us to assemble them with the same end-platform location. In this case it is required to compensate two types of errors (caused by the external loading $\mathbf{F}$ and inaccuracy in the serial chains). The second source of errors can be taken into account by changing of target location $\Delta \mathbf{t}_{0i}$ for each kinematic chain

$$\Delta \mathbf{t}_{0i} = \Delta \mathbf{t}_0 + \Delta \mathbf{t}_\varepsilon - \boldsymbol{\varepsilon}_i \tag{14}$$

where $\Delta \mathbf{t}_\varepsilon$ is the end-platform deflections due to assembling of non-perfect kinematic chains and $\boldsymbol{\varepsilon}_i$ is shifting of the end-point location of $i^{th}$ kinematic chain because of geometrical errors. Using the principle of virtual work it can be proved that $\Delta \mathbf{t}_\varepsilon$ can be computed as

$$\Delta \mathbf{t}_\varepsilon = \left( \sum_{i=1}^{m} \mathbf{K}_C^{(i)} \right)^{-1} \cdot \sum_{i=1}^{m} \left( \mathbf{K}_C^{(i)} \cdot \boldsymbol{\varepsilon}_i \right) \tag{15}$$

where $\mathbf{K}_C^{(i)}$ defines the Cartesian stiffness matrix of $i$-th kinematic chain that can be computed using techniques proposed in the previous Section and $m$ is the number of kinematic chains in the parallel manipulator. A more detailed presentation of the developed iterative routines is given in Figure 3.

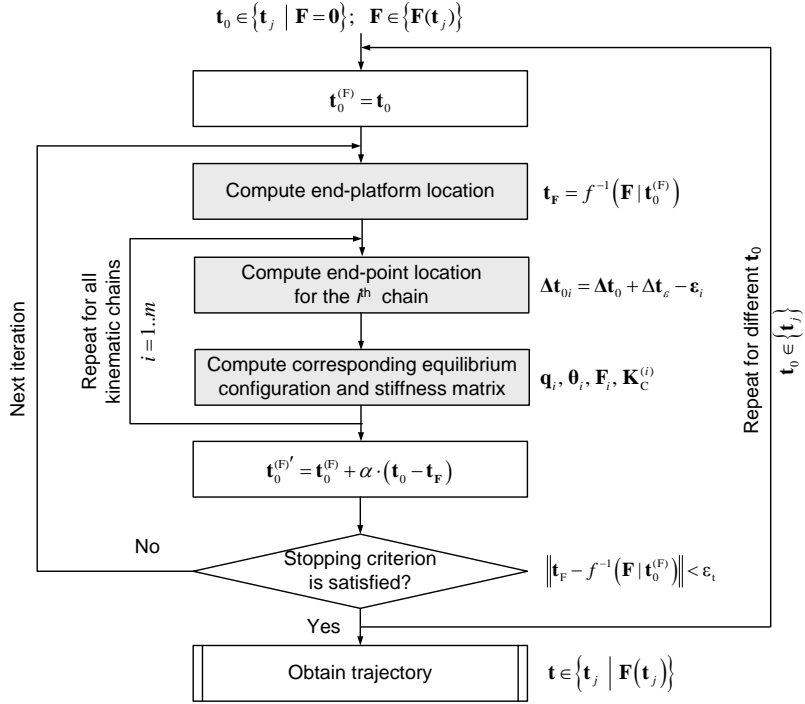

**Figure 3**    Procedure for compensation of compliance errors in parallel manipulator

Hence, using the proposed computational techniques, it is possible to compensate a majority of the compliance errors by proper adjusting the reference trajectory that is used as an input for robotic controller. In this case, the control is based on the inverse kinetostatic model (instead of kinematic one) that takes into account both the manipulator geometry and elastic properties of its links and joints. Efficiency of this technique is confirmed by an example presented in the next section.

## 5  Illustrative example: compliance error compensation for milling

Let us illustrate the developed compliance errors compensation technique by an example of the circle groove milling with the Orthoglide manipulator (Figure 4). Detailed specification of this manipulator can be found in [37]. According to [38], such technological process causes the loading $F_r = 215\,N$ ; $F_t = -10\,N$ ; $F_z = -25\,N$ that together with angular parameter $\varphi = [0, 360°]$ define the forces $F_x, F_y$ and $F_z$ (Figure 4b,c). Here, the tool length $h$ is equal to $100\,mm$. It is assumed that the manipulator has two sources of inaccuracy

  (i)   the assembling errors in the kinematic chains (assembling errors in actuator angular locations of about 1° around the corresponding actuated axis) causing internal forces and relevant deflections in joints and links due to manipulator over-constrained structure;
  (ii)  the external loading $\|\mathbf{F}\| = 217\,N$ caused by the cutting forces, which generates essential compliance deflections causing non-desirable end-platform displacement.

It is worth mentioning that the non-linear compliance error compensation technique, which has been developed in previous section, allows us to compensate influence of both the aforementioned factors

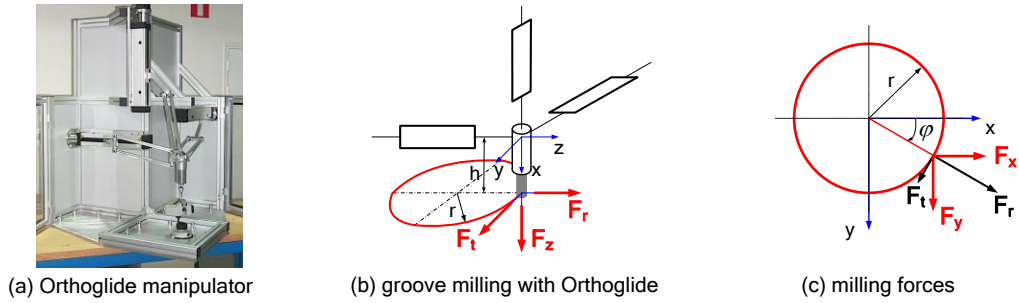

(a) Orthoglide manipulator    (b) groove milling with Orthoglide    (c) milling forces

**Figure 4**   Milling forces for groove milling using Orthoglide manipulator

Assuming that the milling trajectory is oriented in xy-plane, the loading that corresponds to grove milling can be expressed as

$$\mathbf{F} = (F_x,\ F_y,\ F_z, -F_y \cdot h,\ F_x \cdot h,\ 0) \qquad (16)$$

where $F_x$ and $F_y$ depend on the machining tool orientation angle $\varphi$ (Figure 4b,c) as

$$F_x = F_r \cos\varphi + F_t \sin\varphi; \qquad F_y = F_r \sin\varphi + F_t \cos\varphi \qquad (17)$$

In order to illustrate influence of different error sources on the machining trajectory, let us focus on a small radius of the circle that should be machined. In this case, the stiffness matrix is almost the same along the trajectory. Relevant modeling results are presented in Figure 5. They correspond to the vicinity of the Cartesian workspace boundary (i.e., the neighborhood of the point $Q_1$, (126.35mm, 126.35mm, 126.35mm), which is the closest point to the parallel singularity inside of the robot workspace, where the angles between the bar elements of different parallelograms are minimum, see [16] for details). They show the influence of different error sources on the machining trajectory without compensation and the revised machining trajectory that should be implemented in robot controller in order to follow the target trajectory while machining. Here, path 5 compensates the effects seen in path 4 such that circle 1 is achieved. It can be seen that the centre of path 5 is on the opposite side of circle 1 compared to path 4. It can also be seen that the main elliptic direction in path 4 becomes the smallest elliptic direction in path 5. It should be mentioned that because of the torque induced by the cutting forces (tool length 100 mm), the target trajectory and shifted trajectory under the cutting forces are intersecting.

Figure 6 illustrates the superposition principle for the errors caused by an inaccuracy in serial chains and compliance errors caused by cutting forces. The vectors that are used here have been computed for the cases when there is only one source of error (inaccuracy in serial chains or compliance due to cutting forces). The results show that taking into account two error sources simultaneously, the total error is less than the error obtained using superposition principle, but this difference is not high and both trajectories have similar shape and location. The main factor that causes differences between the obtained trajectories is the changes in the stiffness matrix because of changing of the end-point location (the order of errors is not important, in both cases when we take into account the second error it should be computed in the neighborhood of current configuration, however for the results that have been obtained using superposition principle both errors have been computed for the original (target) end-point location).

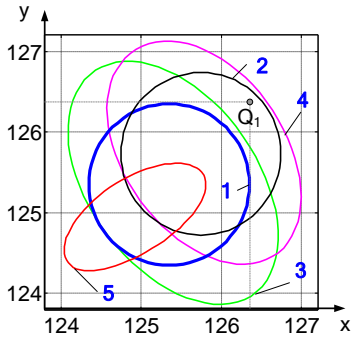

(1) Target trajectory
(2) Shifting of target trajectory caused by errors in serial chains (assembling errors)
(3) Shifting of target trajectory caused by cutting force (compliance errors)
(4) Shifting of target trajectory caused by cutting force and errors in serial chains (assembling errors + compliance errors)
(5) Adjusted trajectory, that insure following the target trajectory while machining

**Figure 5**   Modifications of target trajectory caused by different error sources and adjusted trajectory that insure following the target trajectory while machining

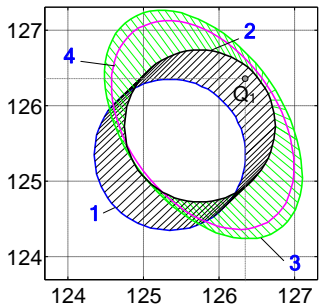

(1) Target trajectory
(2) Shifting of target trajectory caused by errors in serial chains (assembling errors)
(3) Shifting of target trajectory caused by cutting force and errors in serial chains (assembling errors + compliance errors: superposition principle for separate cases)
(4) Shifting of target trajectory caused by cutting force and errors in serial chains (assembling errors + compliance errors: two error sources simultaneously)

**Figure 6**   Superposition of different error sources

Figure 7 presents result for the milling of the 50 mm circle. In this case, without compensation, the compliance errors can exceed 0.8 mm, which is rather high for the considered application. After compensation, the above mentioned errors are reduced near to zero (it is obvious that in practice, the compensation level is limited by the accuracy of the stiffness model). This compensation is achieved due to the modification of the actuator coordinates $\rho$ along the machining trajectory. Compared to the relevant values computed via the inverse kinematics, the actuator coordinates differ up to 1.7 mm. Corresponding forces in actuators can reach 300 N. Some more results on the compliance errors compensation are presented in Figure 7, which includes plots showing modification of the actuator coordinates $\Delta\rho$, values of the compensated end-effector displacement $\Delta t$ and the torques in actuators $\tau$. Figure 7d,e illustrate the impact of different error sources in the inaccuracy while milling with Orthoglide.

Comparison of results for typical locations of the desired circular trajectory are presented in Figure 8. These results include a number of plots showing modification of the actuator coordinates $\Delta\rho$, values of compensated end-effector displacement $\Delta t$ and the torques in actuators $\tau$. It is shown that for such process parameters, without compensation, the compliance errors can exceed 1.2 mm, which is too high for the considered application. In particular, for the best location $Q_0$, the cutting forces provoke the end-effector deflection of 0.35 mm. And for the worst location $Q_3$, the end-effector deflection is about 1.25 mm. Hence, the application of the developed compliance errors compensation technique is reasonable here. Compared to relevant values computed via the inverse kinematics (as in common-used industrial controllers), the actuator coordinates differ up to 0.6 mm for location $Q_0$, and up to 1.9 mm for location $Q_3$.

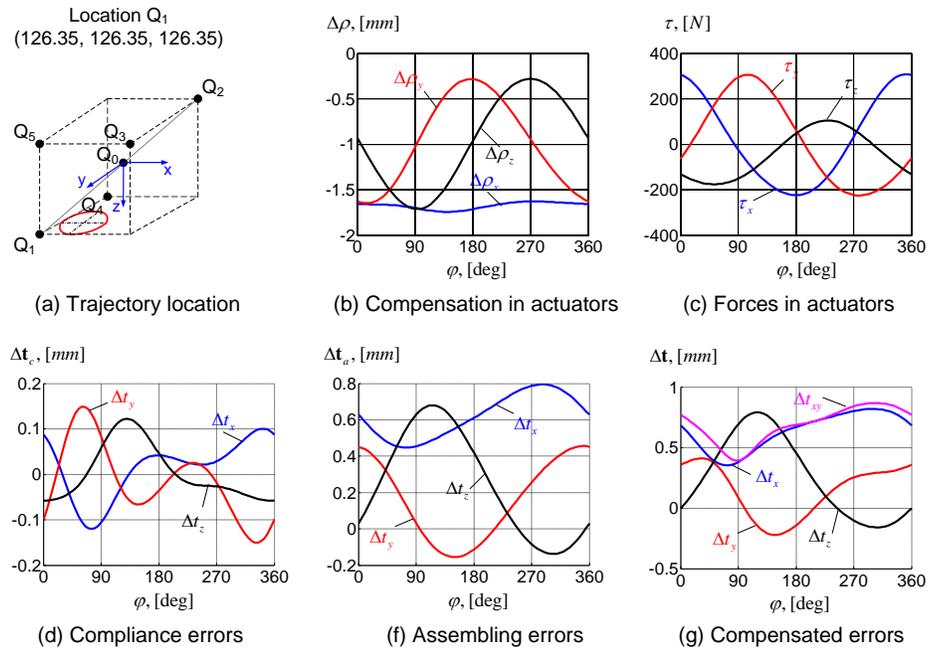

**Figure 7**   Compliance error compensation for Orthoglide milling application

It is worth mentioning, that the shape of the compensation curve $\Delta\rho(\varphi)$ highly depends on the location of the milling trajectory (i.e. the function $\Delta\rho(\varphi)$ cannot be normalized by scaling and shifting) and the compensation procedure requires intensive computing. However, it can be implemented off-line and the robot model motion program can be properly modified.

Hence, the developed algorithm is able to compensate the compliance errors and can be efficient both for off-line trajectory planning and for on-line errors compensation.

## 6   Conclusions

In robotic-based machining, an interaction between the workpiece and technological tool causes essential deflections that significantly decrease the manufacturing accuracy. Relevant compliance errors highly depend on the manipulator configuration and essentially differ throughout the workspace. Their influence is especially important for heavy serial robots and for parallel manipulators, where the compromise between the manipulator stiffness and its dynamic capabilities is quite important. To overcome this difficulty this paper presents a new technique for compensation of the compliance errors caused by external/internal loadings in parallel manipulators (including over-constrained ones) composed of non-perfect serial chains. In contrast to previous works, this technique is based on the non-linear stiffness model.

The advantages of the developed technique are illustrated by the example that deals with groove milling with Orthoglide manipulator. It has been shown that the impact of two error sources cannot be taken into account using their superposition principle due to non-linearity of the stiffness model (even for relatively small deflections of the end-effector). Comparison study confirmed that errors to be compensated highly depend on the workpiece location. Besides, in order to compensate the same error for different workpiece locations different modifications in actuated coordinates are required.

In future, the developed compensation technique will be integrated in a software toolbox. This toolbox will also be useful for optimal path planning as well as optimization of the workpiece location. Another problem that should be addressed is an enhancement of the stiffness modeling technique for a more general class of manipulators and other types of loadings (gravity, friction).

## 7   Acknowledgments

The work presented in this paper was partially funded by the Region "Pays de la Loire", France and by the project ANR COROUSSO, France.

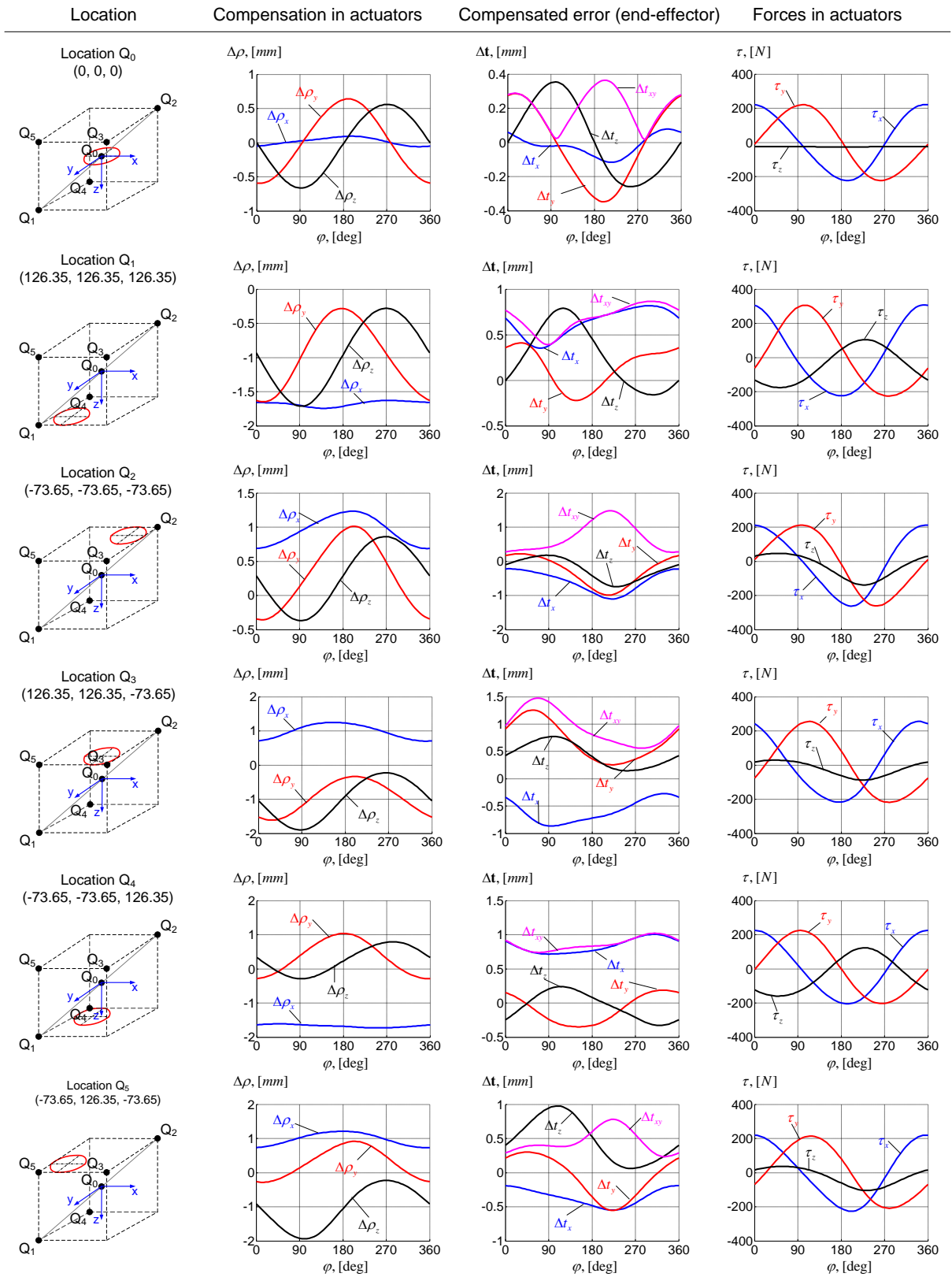

**Figure 8** Compliance error compensation for Orthoglide milling application with cutting force (215 N, -10 N, -25 N, 1 N·m, 21.5 N·m, 0) for different location of the workpiece